\documentclass[a4paper,fleqn]{cas-dc}

\usepackage[numbers]{natbib}
\usepackage{gensymb}

\def\tsc#1{\csdef{#1}{\textsc{\lowercase{#1}}\xspace}}
\tsc{WGM}
\tsc{QE}
\tsc{EP}
\tsc{PMS}
\tsc{BEC}
\tsc{DE}

\begin{document}
\let\WriteBookmarks\relax
\def\floatpagepagefraction{1}
\def\textpagefraction{.001}

\shorttitle{Three-stage binarization of color document images based on discrete wavelet transform and generative adversarial networks}

\shortauthors{Ju~et~al.}

\title [mode = title]{Three-stage binarization of color document images based on discrete wavelet transform and generative adversarial networks}   

\author[1]{Rui-Yang~Ju~}[orcid=0000-0003-2240-1377]
\ead{jryjry1094791442@gmail.com}
\credit{Conceptualization, Methodology, Writing – original draft}

\author[2]{Yu-Shian~Lin~}[orcid=0000-0001-9825-9958]
\ead{abcpp12383@gmail.com}
\credit{Methodology, Data curation, Investigation}

\author[3]{Yanlin~Jin~}[orcid=0000-0001-8466-0660]
\ead{neil.yl.jin@gmail.com}
\credit{Software, Validation, Writing – review \& editing}

\author[2]{Chih-Chia~Chen~}[orcid=0000-0003-0848-9747]
\ead{crystal88irene@gmail.com}
\credit{Formal Analysis, Project administration}

\author[2]{Chun-Tse~Chien~}[orcid=0009-0008-7549-4021]
\ead{popper0927@hotmail.com}
\credit{Visualization}

\author[2]{Jen-Shiun~Chiang~}[orcid=0000-0001-7536-8967]
\cormark[1]
\ead{jsken.chiang@gmail.com}
\credit{Funding acquisition, Supervision, Writing – review \& editing}

\affiliation[1]{
    organization={Graduate Institute of Networking and Multimedia, National Taiwan University},
    addressline={No. 1, Sec. 4, Roosevelt Rd.}, 
    city={Taipei City},
    postcode={106319}, 
    country={Taiwan}}
\affiliation[2]{
    organization={Department of Electrical and Computer Engineering, Tamkang University},
    addressline={No.151, Yingzhuan Rd., Tamsui Dist.}, 
    city={New Taipei City},
    postcode={251301}, 
    country={Taiwan}}
\affiliation[3]{
    organization={Department of Electrical and Computer Engineering, Rice University},
    addressline={6100 Main St.}, 
    city={Houston},
    postcode={77005}, 
    state={Texas},
    country={USA}}

\cortext[cor1]{Corresponding author: jsken.chiang@gmail.com}

\begin{abstract}
The efficient extraction of text information from the background in degraded color document images is an important challenge in the preservation of ancient manuscripts.
The imperfect preservation of ancient manuscripts has led to different types of degradation over time, such as page yellowing, staining, and ink bleeding, seriously affecting the results of document image binarization.
This work proposes an effective three-stage network method to image enhancement and binarization of degraded documents using generative adversarial networks (GANs).
Specifically, in Stage-1, we first split the input images into multiple patches, and then split these patches into four single-channel patch images (gray, red, green, and blue).
Then, three single-channel patch images (red, green, and blue) are processed by the discrete wavelet transform (DWT) with normalization.
In Stage-2, we use four independent generators to separately train GAN models based on the four channels on the processed patch images to extract color foreground information.
Finally, in Stage-3, we train two independent GAN models on the outputs of Stage-2 and the resized original input images ($512 \times 512$) as the local and global predictions to obtain the final outputs.
The experimental results show that the Avg-Score metrics of the proposed method are 77.64, 77.95, 79.05, 76.38, 75.34, and 77.00 on the (H)-DIBCO 2011, 2013, 2014, 2016, 2017, and 2018 datasets, which are at the state-of-the-art level.
The implementation code for this work is available at \url{https://github.com/abcpp12383/ThreeStageBinarization}.
\end{abstract}


\begin{keywords}
Deep learning \sep
Computer vision \sep
Discrete wavelet transform \sep
Generative adversarial networks \sep
Document image processing \sep
Document image enhancement \sep
Document image binarization
\end{keywords}

\maketitle
\section{Introduction}
Historical analysis of ancient human society mainly relies on the reconstruction and analysis of preserved ancient documents, where the parchment manuscripts are an important tool for preserving important ancient documents.
Analyzing preserved ancient documents, especially the parchment manuscripts, presents significant challenges due to potential forms of degradation, including paper yellowing, text fading, page contamination, and moisture damage~\cite{hedjam2013historical,sun2016blind}.

Document image enhancement and binarization are important steps in analyzing ancient documents, with the purpose of extracting useful text information from the background.
However, improving the image quality of degraded documents requires solving several problems, such as text degradation and ink bleeding~\cite{kligler2018document,sulaiman2019degraded}.
Traditional threshold methods cannot effectively extract text information from the background of degraded documents, often resulting in incomplete removal of shadows and noise, or loss of text information~\cite{niblack1985introduction,otsu1979threshold,sauvola2000adaptive}.

With the development of deep learning, neural networks have been applied to image binarization.
The common method is to apply neural networks to analyze the document images, for instance, Calvo-Zaragoza and Gallego~\cite{calvo2019selectional} introduce a selectional auto-encoder approach to analyze document images through the trained network model, and binarize the document images using the threshold.

The classic network in semantic segmentation tasks, fully convolutional network (FCN)~\cite{long2015fully}, can also be used to segment foreground and background information in document images.
Tensmeyer \emph{et al.}'s~\cite{tensmeyer2017document} research also shows that using FCN for document binarization can achieve good results.
However, most current methods mainly process gray-scale document images, targeting contaminated black and white scanned ancient documents as their default.

Considering some ancient document images are in color, we propose a novel three-stage network method for both gray-scale and color degraded document image enhancement and binarization.
Specifically, the proposed method consists of three stages.
In Stage-1, we first split the input images into multiple patches, and then split these patches into four single-channel patch images (gray, red, green, and blue).
After that, we apply the discrete wavelet transform (DWT) on three single-channel patch images (red, green, and blue) to retain the low-low (LL) subband images, and then normalize them.
To remove the noise from the background of the original input images, and to extract the foreground information from the patch images, in Stage-2, we train the GANs model on four single-channel images by using four independent generators to generate the enhanced images.
In addition, in Stage-3, we use the entire document images composed of the patch images (output of Stage-2) as the local predictions, and the $512 \times 512$ resized input images as the global predictions to train two independent GAN models, respectively.
The outputs of these two GAN models are fused to obtain the final output images.

To evaluate the performance of the proposed method for document image binarization, we compare it with other methods using the (H)-DIBCO datasets of 2011, 2013, 2014, 2016, 2017, and 2018~\cite{pratikakis2011icdar2011,pratikakis2013icdar,ntirogiannis2014icfhr2014,pratikakis2016icfhr2016,pratikakis2017icdar2017,pratikakis2018icdar2018}.

The main contributions of this paper are as follows:
\begin{itemize}
\item Introduces an effective three-stage network method that effectively performs enhancement and binarization on gray-scale and color degraded document images.
\item Shows that, compared with other methods using the same training set, the proposed method achieves the state-of-the-art performance on multiple benchmark datasets.
\item Demonstrates that for document images, the binarization results of the images processed by the discrete wavelet transform with normalization are closer to the Ground-Truth images compared to the binarization results of the original input images.
\end{itemize}

The rest of this paper is organized as follows:
Section~\ref{sec:related} describes the deep learning methods for document image enhancement and binarization.
Section~\ref{sec:method} details the proposed three-stage network method.
Section~\ref{sec:experiments} presents the results of the image enhancement experiment and the comparison experiment, and shows the comparative results of the performance of the proposed method and other methods on multiple benchmark datasets.
Finally, Section~\ref{sec:conclusion} discusses the conclusion and future work of this work.

\section{Related Work}
\label{sec:related}
Document image binarization methods can be broadly divided into two categories: traditional threshold methods and deep learning methods.
Among them, traditional threshold methods are mainly based on calculating the threshold value at the pixel-level of the image, such as global binarization~\cite{otsu1979threshold} and local adaptive binarization~\cite{niblack1985introduction,sauvola2000adaptive}.
However, these methods have unsatisfactory results for documents affected by excessive interference or degradation~\cite{howe2013document,jia2018degraded}.

Neural networks~\cite{zeiler2014visualizing,krizhevsky2017imagenet} continue to achieve excellent results in computer vision tasks.
Among them, neural networks used in semantic segmentation tasks are also applied to document image binarization.
For instance, Tensmeyer \emph{et al.}~\cite{tensmeyer2017document} apply FCN~\cite{long2015fully} to document image binarization, and combine pseudo-F-measure (p-FM) and F-measure (FM) losses to train the model.
Inspired by U-Net~\cite{ronneberger2015u}, Peng \emph{et al.}~\cite{peng2017using} introduce an encoder-decoder architecture for document image binarization, where the decoder transforms the low-resolution representation to the original dimensions, resulting in the binarized images.
Vo \emph{et al.}~\cite{vo2018binarization} apply hierarchical deep supervised network (DSN) to document image binarization, which achieves the state-of-the-art (SOTA) performance on multiple benchmark datasets.
He \emph{et al.}~\cite{he2019deepotsu} propose an iterative deep learning framework applied to document image binarization, which also achieves the excellent performance.

In recent years, GANs~\cite{goodfellow2020generative} have achieved excellent results in image generation tasks.
Different from generating output images from random noise, Isola \emph{et al.}~\cite{isola2017image} propose Pix2Pix GAN based on conditional generative adversarial network (cGAN)~\cite{mirza2014conditional} for image-to-image translation.
To ensure that the generated images are more realistic, Pix2Pix GAN employs generators to optimize the loss function.
In the network architecture, the generator evaluates the minimization of generated images against the target output images, quantifying the $L1$ loss between them.
Meanwhile, the discriminator evaluates the credibility of the generated images based on the input images and the reference target images (the Ground-Truth images).
The process is described by the following equation:

\begin{equation}
\label{eq:pix2pix}
\begin{split}
\emph{arg} \min_{G} \max_{D} V(G,D) = \mathop{\mathbb{E}_{x}}[\log (1-D(x, G(x)))] \\
+ \mathop{\mathbb{E}_{x,y}}[\log D(x,y)] + \lambda \mathop{\mathbb{E}_{x,y}}[|| y - G(x) ||_1],
\end{split}
\end{equation}
where the input data distribution $\mathop{\mathbb{P}_x}$ samples the input images $x$, and the Ground-Truth images are $y$.
The hyperparameter to increase the model regularization effect is $\lambda$.
Generator ($G$) generates the input images as the generated images $G(x)$.
From Eq.~(\ref{eq:pix2pix}), the loss function between the generated images and the Ground-Truth images is $L1$ instead of $L2$, which is beneficial to reduce blurring and ambiguity in the generation process.

Therefore, in addition to neural networks applied in semantic segmentation tasks, GANs have also achieved satisfactory results for document image binarization.
Bhunia \emph{et al.}~\cite{bhunia2019improving} employ cGAN \cite{mirza2014conditional} on the DIBCO datasets for document image enhancement and binarization. 
Inspired by Pix2Pix GAN~\cite{isola2017image}, Zhao \emph{et al.}~\cite{zhao2019document} propose a cascade generator structure for document image binarization based on Pix2Pix GAN to solve the problem of multi-scale information combination.
De \emph{et al.}~\cite{de2020document} design two discriminators to combine local and global information.
Suh \emph{et al.}~\cite{suh2022two} propose a novel two-stage generative adversarial network method using PatchGANs~\cite{isola2017image} for document image enhancement and binarization.
Souibgui \emph{et al.}~\cite{souibgui2020gan} introduce an efficient end-to-end framework named document enhancement generative adversarial networks (DE-GAN).
This framework is designed to recover severely degraded document images by generating high-quality enhanced versions of these documents.
These methods keep breaking the SOTA performance on the DIBCO datasets.

In addition, diffusion models have also begun to be applied to document image binarization.
For instance, Cicchetti and Comminiello~\cite{cicchetti2024naf} propose a novel generative framework based on the diffusion probabilistic model (DPM) for document image enhancement, which includes an efficient nonlinear activation free (NAF) network and an additional differentiable module based on the convolutional recurrent neural network (CRNN).

\begin{figure*}[ht]
\centering
\includegraphics[width=\linewidth]{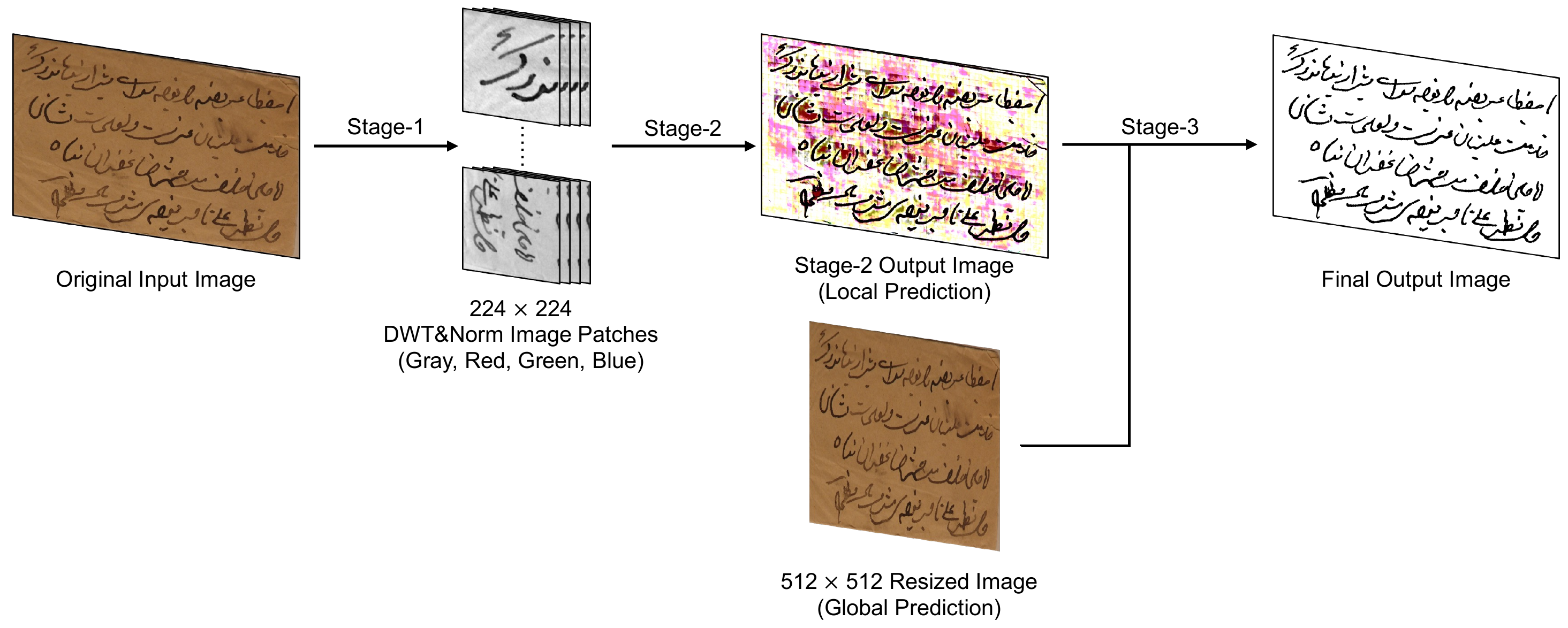}
\caption{The overall three-stage network architecture of the proposed method for document image enhancement and binarization.}
\label{figure_overall}
\end{figure*}

\begin{figure*}[ht]
\centering
\includegraphics[width=\linewidth]{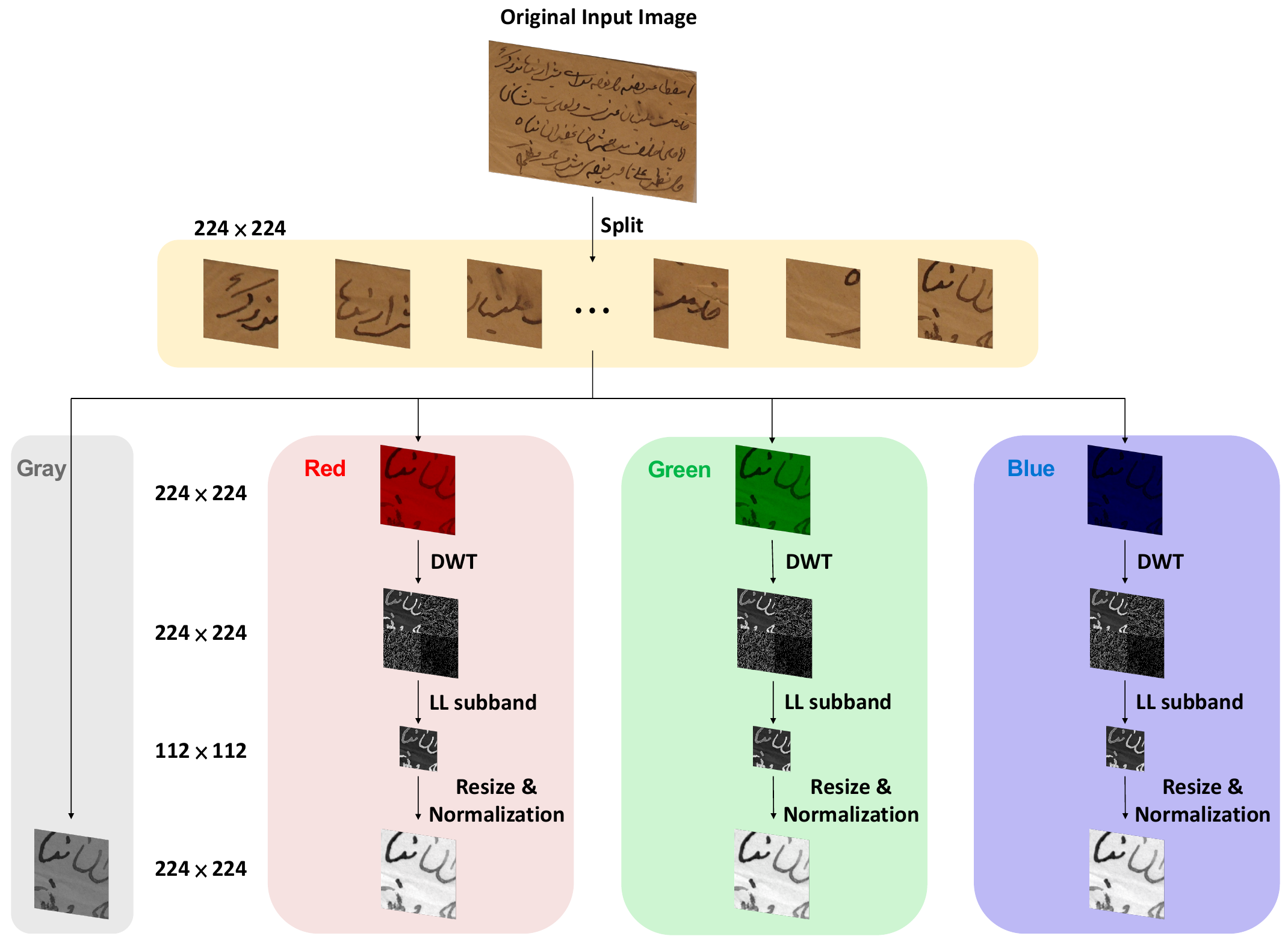}
\caption{The architecture of Stage-1 of the proposed method.
The original input images are split into multiple patches ($224 \times 224$), and split into four single-channel patch images (gray, red, green, and blue), where three single-channel patch images (red, green, and blue) are processed by the Haar wavelet transform with normalization.}
\label{figure_stage_1}
\end{figure*}

\begin{figure*}[ht]
\centering
\includegraphics[width=\linewidth]{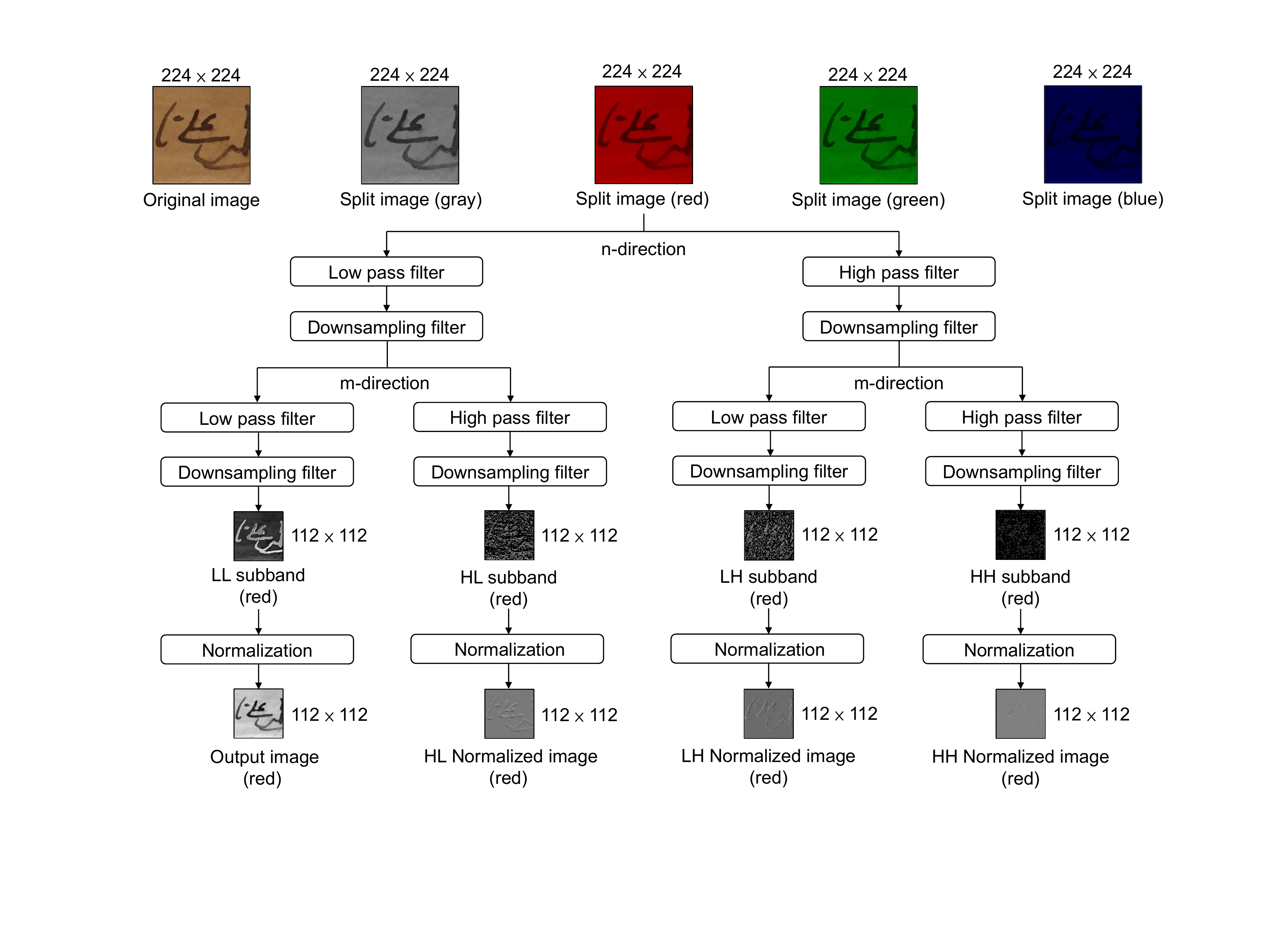}
\caption{The flow chart of the Haar wavelet transform with normalization used by the proposed method, which takes the red single-channel patch images as the example.
The processes of the green and blue single-channel patch images are the same.}
\label{figure_DWT}
\end{figure*}

\begin{figure*}[ht]
\centering
\includegraphics[width=\linewidth]{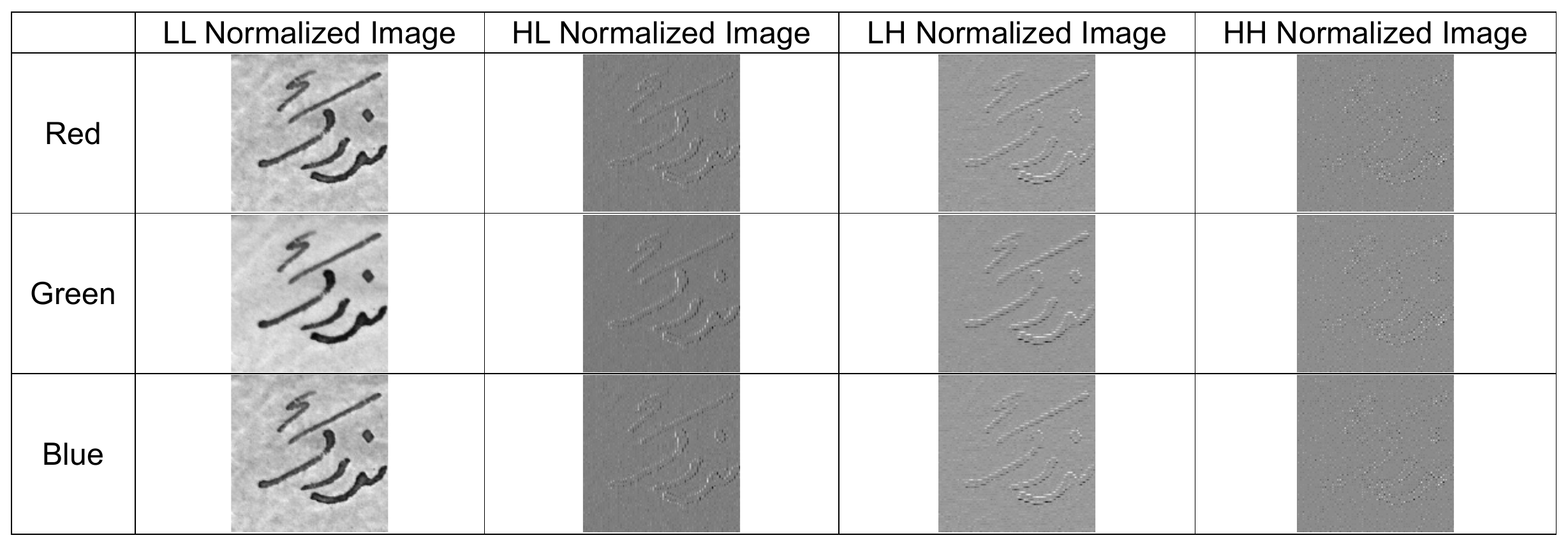}
\caption{Example of the Haar wavelet transform with normalization of three single-channel patch images (red, green, and blue).}
\label{figure_color}
\end{figure*}

\begin{figure*}[ht]
\centering
\includegraphics[width=\linewidth]{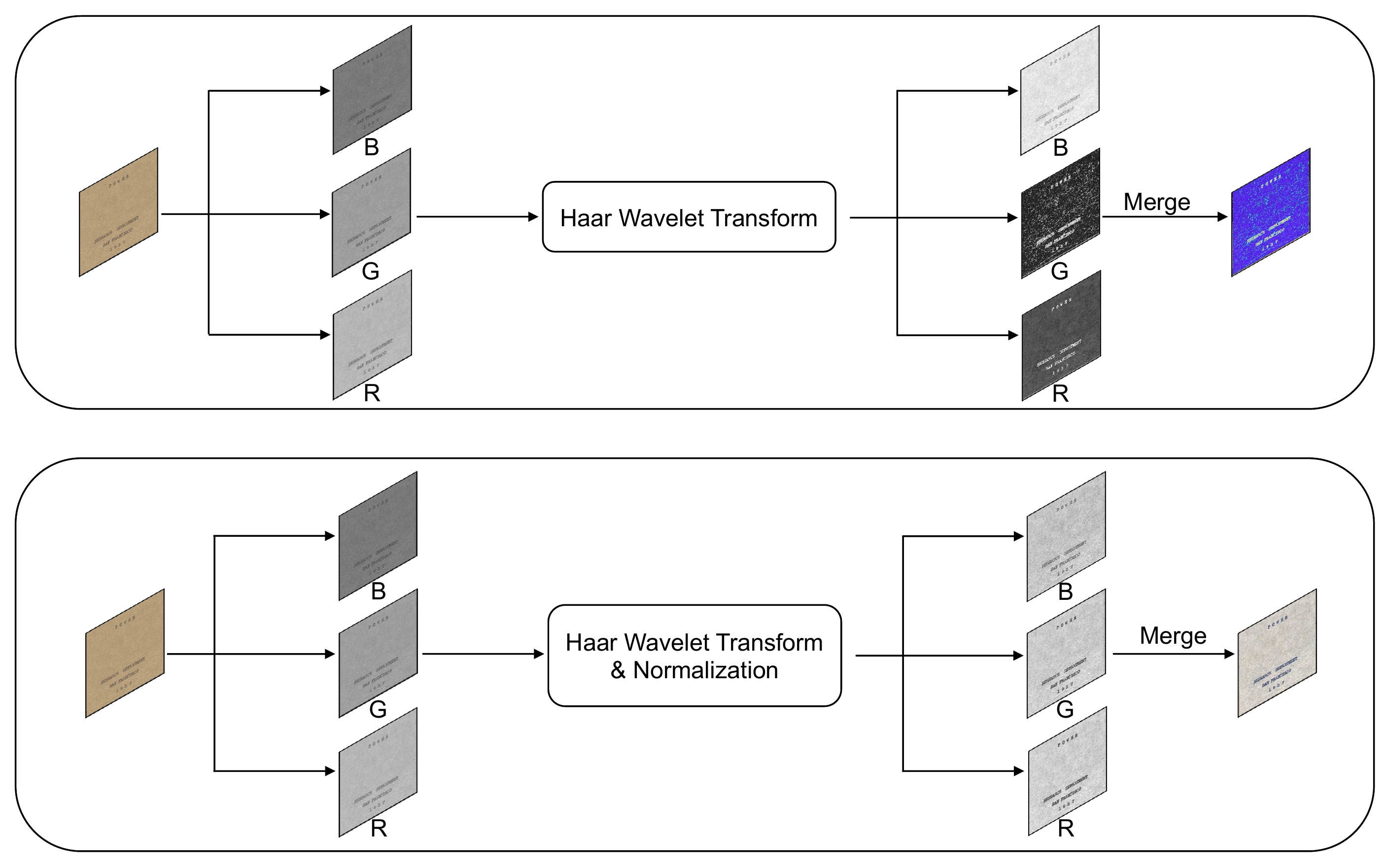}
\caption{Example of the ablation study on image normalization, the above figure presents the image only processed by the Haar wavelet transform, and the below figure presents the image processed by the the Haar wavelet transform with normalization.}
\label{figure_ablation}
\end{figure*}

\begin{figure*}[ht]
\centering
\includegraphics[width=\linewidth]{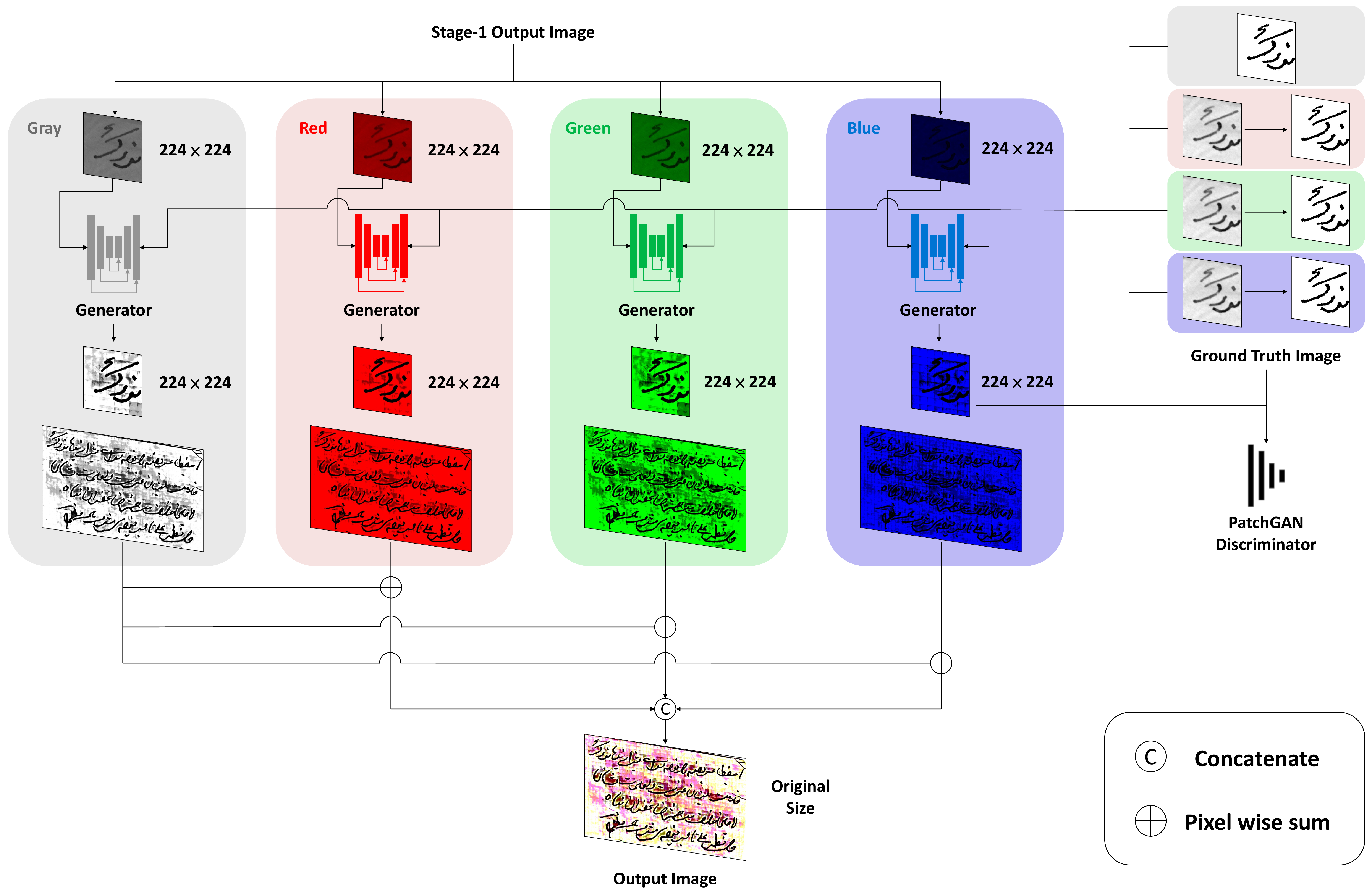}
\caption{The architecture of Stage-2 of the proposed method.
This method employs the U-Net++~\cite{zhou2018unet++,zhou2019unet++} architecture with EfficientNet-B6~\cite{tan2019efficientnet} as the generator, and the improved PatchGAN~\cite{isola2017image} as the discriminator.
The Ground-Truth images for three single-channel patch images (red, green, and blue) are the binarization results of the outputs from Stage-1.}
\label{figure_stage_2}
\end{figure*}

\begin{figure*}[ht]
\centering
\includegraphics[width=\linewidth]{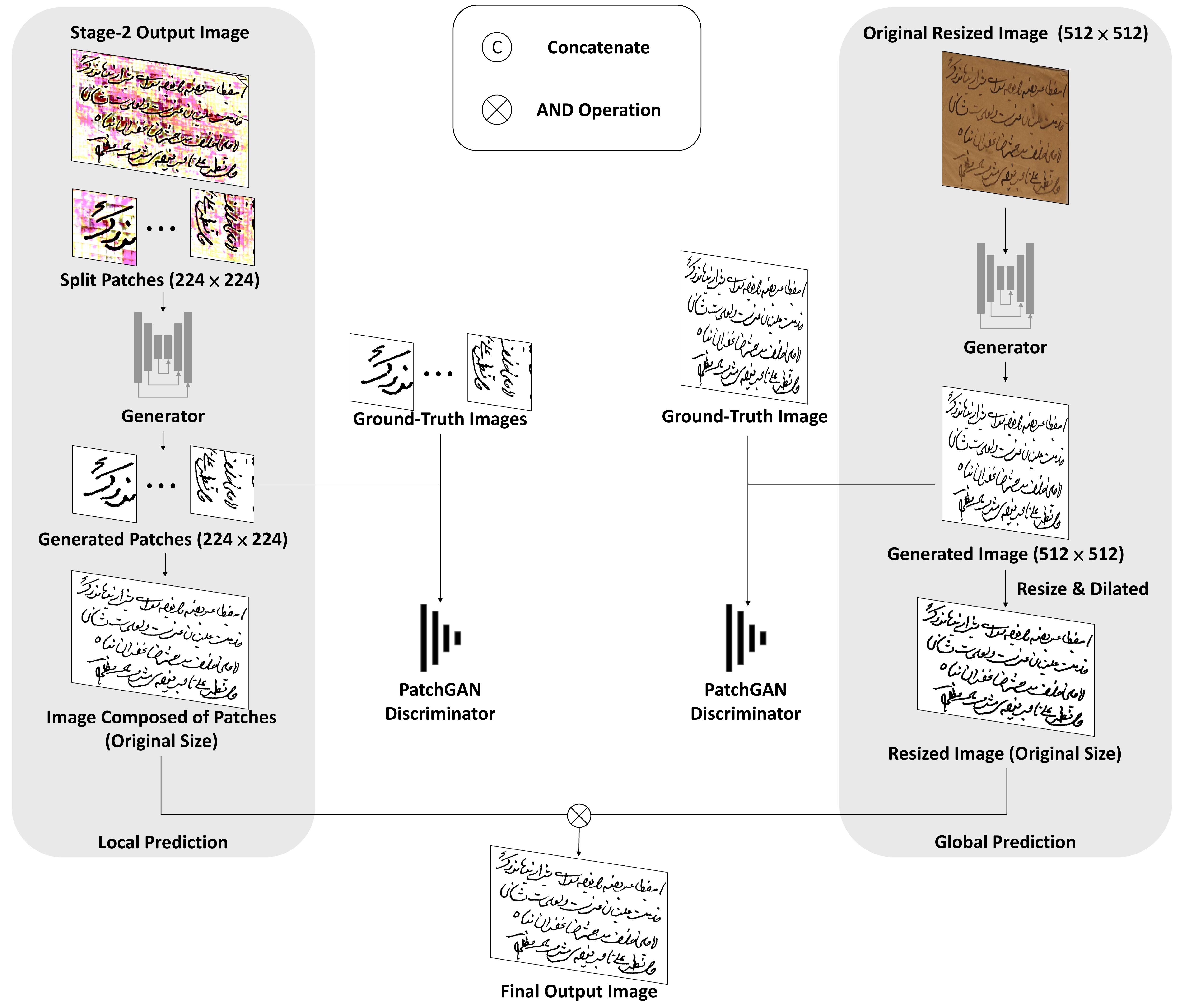}
\caption{The architecture of Stage-3 of the proposed method.
This stage comprises the local prediction (left-side) and the global prediction (right-side).}
\label{figure_stage_3}
\end{figure*}

\section{Proposed Method}
\label{sec:method}
\subsection{Overall Network Architecture}
Considering the complexity of degradation of color documents, better results are often obtained by processing images of different color channels separately~\cite{suh2022two}.
Therefore, this method divides the overall network architecture into three stages, which are used for document image enhancement, text information extraction, and document image binarization, as shown in Figure~\ref{figure_overall}.
The images processed in Stage-1 are prepared for training GANs models in Stage-2 and Stage-3.
Stage-2 focuses on the removal of noise from the background and the extraction color foreground information, while Stage-3 involves the fusion of the local and global predictions to obtain the final output images.
The detailed introduction of each stage can be found in Sections~\ref{sub:stage-1},~\ref{sub:stage-2}, and~\ref{sub:stage-3}, respectively.

\subsection{Stage-1}
\label{sub:stage-1}
This work employs four independent generators to train GANs model, which extracts foreground information from degraded document images while removing noise from the background.
For color degraded documents, to process the inputs of different independent generators, we first split the original input images (24-bit) into four single-channel patch images (gray, red, green, and blue), each of which is 8-bit.
However, for the gray-scale (8-bit) degraded documents, we only process the gray single-channel images in Stage-1 and train the model on gray single-channel images for Stage-2.

Next, we will focus on the case of color degraded documents.
Since the Ground-Truth images provided by the dataset are binarized images (8-bit), images with different channel numbers cannot be directly put into the discriminator for processing.
As shown in Figure~\ref{figure_stage_1}, except for the gray single-channel images, we process the remaining three single-channel images (red, green, and blue) using the Haar wavelet transform (one of the DWTs), and retain the LL subband images for normalization.
Since the sizes of the retained LL subband images are $112 \times 112$ (half of $224 \times 224$), we resize the images to the original size of $224 \times 224$ using the bicubic interpolation~\cite{keys1981cubic}.

In addition, as illustrated in Figure~\ref{figure_stage_1}, the Ground-Truth images used for the four channels are different.
For the gray channel, we directly use the Ground-Truth images provided by the dataset.
For the remaining three channels (red, green, and blue), we employ the results of global binarization of the single-channel images processed by the DWT with normalization as the Ground-Truth images.
The reasons for using DWT with normalization and their effects on the images are described in Section~\ref{subsec:experiment_norm}.

\subsubsection{Discrete Wavelet Transform}
In this work, we use the Haar wavelet transform~\cite{haar1909theorie} (one of the DWTs) to process the input images.
This processing can be equated to using a bandpass filter that selectively allows signals with frequencies similar to the wavelet basis functions to pass through.
The flowchart of this processing is illustrated in Figure~\ref{figure_DWT}, which generates four subband images, such as the low-low (LL) subband images, the high-low (HL) subband images, the low-high (LH) subband images, and the high-high (HH) subband images.
The low-pass filtered signal $V_L$ at position $(m,n)$ is obtained by convolving the input signal $x$ with the low-pass filter $g$, where the input signal is downsampled by a factor of 2, as shown in the following equation:
\begin{equation}
\begin{split}
v_{L}[m,n]=\sum_{k=0}^{K-1}x[m,2n-k]g[k]
\end{split}
\label{eq:1-D_DWT_Low}
\end{equation}
where $x[m,2n-k]$ is the input signal $x$ at position $(m,2n-k)$, and this indexing suggests downsampling by a factor of 2. $g[k]$ is the filter coefficients for the low-pass filter, and $K$ is the length of the filter $g$.
The low-pass filtered signal $V_LL$ at position $(m,n)$ is obtained by convolving the intermediate low-pass filtered signal $V_L$ (filtered along the columns) with the low-pass filter $g$, where the intermediate signal is downsampled by a factor of 2 along the row dimension, as shown in the following equation:
\begin{equation}
\begin{split}
v_{LL}[m,n]=\sum_{k=0}^{K-1}v_{L}[2m-k,n]g[k]	
\end{split}
\end{equation}

In contrast to the Ground-Truth images provided by the dataset, the three single-channel images (red, green, and blue) possess a significant amount of noise.
The noise will prevent them from being used directly through the generator to generate the corresponding Ground-Truth images and possibly severely affect the performance of the trained GANs models.
The DWT decomposes the images into both low-frequency and high-frequency information, where the low-frequency information represents the mean value, and the high-frequency information represents the difference value.
The mean value captures gradients and encodes both contour information and proximate details of the images, and the difference value captures rapid changes and encodes complex image details and local information.
Therefore, we apply the DWT and retain the LL subband images, as shown in Figure~\ref{figure_DWT}, which effectively preserves contour information and reduces the noise interference.

\subsubsection{Normalization}
Normalization entails constraining the data to the desired range with the aim of simplifying data processing and speeding up convergence.
The proposed equation for this process is expressed as:
\begin{equation}
\begin{split}
I_{N}=(\operatorname{IV_{nmax}}-\operatorname{IV_{nmin}}) \frac{1}{1+e^{-\frac{I-\beta}{\alpha}}}+\operatorname{IV_{nmin}}
\end{split}
\end{equation}
where $IV$ is the intensity values, the image to be processed $I:\left\{\mathbb{X} \subseteq \mathbb{R}^{n}\right\} \rightarrow\{\operatorname{IV_{min}}, \ldots, \operatorname{IV_{max}}\}$, and the generated image $I_N:\left\{\mathbb{X} \subseteq \mathbb{R}^{n}\right\} \rightarrow\{\operatorname{IV_{nmin}}, \ldots, \operatorname{IV_{nmax}}\}$.
$\alpha$ defines the width of the input intensity range, and $\beta$ defines the range centered intensity.

In our method, we normalize the LL subband images by the DWT to present initially incomparable data for comparison.
The reason we choose the normalized LL subband images as the output of Stage-1 is shown in Figure~\ref{figure_color}, where the foreground information in another subband images is not prominent compared to the LL subband images.

Figure~\ref{figure_ablation} illustrates the example of the effect of normalization on three single-channel patch images (red, green, and blue) processed by the DWT, which effectively balances three colors when applied to the ``HW15'' image of the DIBCO 2011 dataset~\cite{pratikakis2011icdar2011}.
In contrast, the output images without normalization are heavily biased towards the blue component.
In conclusion, the main difference between the results of Figure~\ref{figure_ablation} (a) and (b) is that the images without normalization may lead to noise in the single-channel images, which seriously affect the pixel-level summation results.

In addition, we also show that, compared to the binarization results of the images only processed by the DWT, the binarization results of the images using the DWT with normalization are more similar to the Ground-Truth images provided by the dataset.
The related experiments are explained in detail in Section~\ref{subsec:experiment_norm}.

\subsection{Stage-2}
\label{sub:stage-2}
In Stage-2, the proposed method is illustrated in Figure~\ref{figure_stage_2}.
Each single-channel patch image is input into an independent generator, so four generators are used in Stage-2 of the proposed method, which enable the method to extract color foreground information while removing background noise from the patch images.
In addition, all single-channel images share the same discriminator to distinguish between the generated images and the corresponding Ground-Truth images.

\subsubsection{Loss functions of GAN}
To ensure a more robust convergence of the loss, we follow an improved loss function of the Wasserstein generative adversarial network with gradient penalty (WGAN-GP)~\cite{gulrajani2017improved} used by Suh \emph{et al.}~\cite{suh2022two}, where WGAN-GP loss incorporates the BCE loss.
Since Bartusiak \emph{et al.}'s~\cite{bartusiak2019splicing} experiments demonstrate that the binary cross-entropy (BCE) loss outperforms the $L1$ loss in binary classification tasks, Suh \emph{et al.}'s~\cite{suh2022two} loss function employs the BCE loss instead of the $L1$ loss used in the Pix2Pix GAN~\cite{isola2017image}.
This improved loss function of the WGAN-GP is as follows:
\begin{equation}
\label{eq:theta_G}
\begin{split}
\mathop{\mathbb{L}_G} = \mathop{\mathbb{E}_{x}}[D(G(x),x)] + \lambda \mathop{\mathbb{E}_{G(x),y}} [y\log G(x) \\
+ (1-y)\log (1-G(x))]	
\end{split}
\end{equation}

\begin{equation}
\label{eq:theta_D}
\begin{split}
\mathop{\mathbb{L}_D} = - \mathop{\mathbb{E}_{x,y}}[D(y,x)] + \mathop{\mathbb{E}_{x}}[D(G(x), x)] \\
+ \alpha \mathop{\mathbb{E}_{x, \hat{y}\sim P_{\hat{y}}}}[( || \nabla_{\hat{y}} D(\hat{y}, x) ||_2 -1 )^2]
\end{split}
\end{equation}
where the penalty coefficient is $\alpha$, and the uniform sampling along a straight line between the Ground-Truth distribution $P_y$, and the point pairs of the generated data distribution is $P_{\hat{y}}$.
$\lambda$ is used to control the relative importance of different loss terms.
In the generator, the distances between the generated images and the Ground-Truth images in each color channel are minimized by the loss function $\mathbb{L}_G$ in Eq.~(\ref{eq:theta_G}).
In the discriminator, the generated images are distinguished from the real images by the loss function $\mathbb{L}_D$ in Eq.~(\ref{eq:theta_D}).

\begin{algorithm}
\caption{Training process for the proposed GAN models}
\begin{algorithmic}[1]
\REQUIRE Input images $I_{input}$
\ENSURE $\omega = 0.5$, $\alpha = 10$, $\lambda=50$, $\eta = 1\times 10^{-4}$
\STATE \textbf{Initialize:} Parameters $\theta_{G_r}$, $\theta_{G_g}$, $\theta_{G_b}$, $\theta_{G_{gray}}$
\FOR{number of training iterations}
\STATE Split $I_{input}$ into $\{red, green, blue, gray\}$ images
\FOR{$k=\{red, green, blue, gray\}$}
\IF{$k$ is gray}
\STATE $y_k \gets y$
\ELSE
\STATE $x'_k=Norm(V_{LL}(x_k))$
\STATE $y_k \gets x'_k \bigcap y$
\STATE Binarize $y_k$ with threshold value $t$.
\ENDIF
\STATE Update Discriminator $D$ (Eq. (\ref{eq:theta_D})):
\STATE $\theta_D \gets \theta_D - \eta_D \nabla_{\theta_D} \mathop{\mathbb{L}_D}$
\STATE Update Generator $G_k$ (Eq. (\ref{eq:theta_G})):
\STATE $\theta_{G_k} \gets \theta_{G_k} - \eta_G \nabla_{\theta_{G_k}} \mathop{\mathbb{L}_G}$
\ENDFOR
\ENDFOR
\FOR{$k=\{red, green, blue\}$}
\STATE $\hat{y}_k \gets \omega G_k(x_k) + (1-\omega) G_{gray}(x_{gray})$
\ENDFOR
\STATE $\hat{y} \gets [\hat{y}_{r}, \hat{y}_{g}, \hat{y}_{b}]$
\end{algorithmic}
\label{algorithm}
\end{algorithm}

\subsubsection{Network architecture}
For the network architecture proposed in Stage-2, the four independent generators use the same encoder-decoder architecture, where the encoder extracts the features and multiple encoders of different depths use a shared decoder and subsequently concatenate the features.
A skip connection is used between the encoder and the decoders.
The encoder facilitates downsampling and context extraction, while the decoders undertake upsampling and amalgamate both upscaled and downsampled features.
Specifically, we employ EfficientNet-B6~\cite{tan2019efficientnet} as the encoder, which is a lightweight convolutional neural network known for its performance in image classification tasks.
Instead of U-Net~\cite{ronneberger2015u} used by Suh \emph{et al.}~\cite{suh2022two}, we use the improved model U-Net++~\cite{zhou2018unet++,zhou2019unet++} as the network architecture for the generator, which can further improve the model performance.

In addition, we follow the discriminator used by Suh \emph{et al.}~\cite{suh2022two} as the discriminator in our network architecture due to its better generalization ~\cite{isola2017image,zhao2019document,de2020document,li2016precomputed,zhu2017unpaired,ledig2017photo}.

The detailed procedure for training the proposed GANs models is illustrated in Algorithm~\ref{algorithm}.
The training process is based on our constructed training environment, which includes the optimizer selection, learning rate, and loss function settings, as described in Section~\ref{sec:implementation}.
The algorithm demonstrates that the GANs models iteratively update the loss function based on the discriminator's assessment of the trueness of the generated images.
This iterative process enhances the generator's ability to generate more realistic images.

\subsection{Stage-3}
\label{sub:stage-3}
In Stage-3, we employ multi-scale GANs models to generate images for both local and global binarization results to enhance the distinction between foreground and background with higher accuracy.
In contrast to Stage-2, which relies on the local prediction using patch images ($224 \times 224$), Stage-3 performs the global prediction on the resized original input images ($512 \times 512$).
Therefore, this method reduces the contextual information loss in text caused by the local predictions of Stage-2. 

Figure~\ref{figure_stage_3} illustrates the process of two independent discriminators in Stage-3. 
Different from Stage-2 using 8-bit images for the local prediction, Stage-3 uses 24-bit images as the inputs of the generator.
In addition, Figure~\ref{figure_stage_3} illustrates that the left-side input images consist of the output images from Stage-2, which are used for the local prediction.
In contrast, the right-side input images represent the resized images ($512 \times 512$) from the original input images for the global prediction.
It is worth noting that in the binarization process of Stage-3, the network architecture of the generator remains the same as in Stage-2, differing only in the number of input channels.

\section{Experiments}
\label{sec:experiments}
\subsection{Datasets}
\subsubsection{DIBCO Dataset}
The document image binarization competition (DIBCO) provides nine datasets~\cite{gatos2009icdar,pratikakis2010h,pratikakis2011icdar2011,pratikakis2012icfhr,pratikakis2013icdar,ntirogiannis2014icfhr2014,pratikakis2016icfhr2016,pratikakis2017icdar2017,pratikakis2018icdar2018}: DIBCO 2009, H-DIBCO 2010, DIBCO 2011, H-DIBCO 2012, DIBCO 2013, H-DIBCO 2014, H-DIBCO 2016, DIBCO 2017, and H-DIBCO 2018.
These datasets comprise both gray-scale and color images, including machine-printed as well as handwritten content.
Participants in the competition were tasked with designing algorithms capable of extracting binarized text images from both handwritten and machine-printed images within these datasets.

\subsubsection{PHIBD}
The persian heritage image binarization dataset (PHIBD)~\cite{ayatollahi2013persian} contains 15 images of historical manuscripts from Mirza Mohammad Kazemaini's old manuscript library in Yazd, Iran.
The images within the dataset have suffered from various types of degradation. 

\subsubsection{SMADI}
The synchromedia multispectral ancient document images (SMADI) dataset~\cite{hedjam2013historical} consists of 240 multispectral images of 30 authentic historical handwritten letters.
These letters, dating from the 17th to the 20th centuries, are written using iron gall ink.
The original documents are deposited at the BAnQ Bibliotheque et Archives nationales du Quebec. 
A total of 240 images of real documents are meticulously captured, calibrated, and registered within this dataset.
Each of these images is imaged with a CROMA CX MSI camera, resulting in 8 images per document.

\subsubsection{Bickley Diary Dataset}
The bickley diary (BD) dataset~\cite{deng2010binarizationshop} has been generously donated to the Singapore Methodist Archives by Erin Bickley.
This dataset comprises diaries in which the images have been affected by both light and fold damages, thereby complicating the process of identification.

\subsection{Evaluation Metric}
\label{subsec:metric}
In comparison with other methods for document image enhancement and binarization, based on the case that DIBCO datasets do not provide OCR output, we use the following five metrics:
\subsubsection{F-measure (FM)}
\begin{equation}
\label{eq:FM}
{\rm FM} = \frac{2 \times {\rm Recall} \times {\rm Precision}}{{\rm Recall} + {\rm Precision}}
\end{equation}
where the $Recall$ is defined as $\frac{TP}{TP + FN}$; the $Precision$ is defined as $\frac{TP}{TP + FP}$, and TP, FP, and FN represent the true positive, false positive, and false negative values.

\subsubsection{Pseudo-F-measure (p-FM)}
\begin{equation}
{\rm p-FM} = \frac{2 \times {\rm Recall_{ps}} \times {\rm Precision_{ps}}}{{\rm Recall_{ps}} + {\rm Precision_{ps}}}
\end{equation}
where both pseudo-recall (${\rm Recall_{ps}}$) and pseudo-precision (${\rm Precision_{ps}}$) use distance weights relative to the contours of the Ground-Truth images to generate new weighted images.
    
\subsubsection{Peak signal-to-noise ratio (PSNR)}
\begin{equation}
\label{eq:psnr}
{\rm PSNR} = 10 \log (\frac{{\rm V}^2}{{\rm MSE}})
\end{equation}
where ${\rm V}$ denotes the difference between the foreground and background, and the mean square error (MSE) is defined as:
\begin{equation}
{\rm MSE} = \frac{\sum_{x=0}^{M-1}\sum_{y=0}^{N-1}{[}L(x,y)-L'(x,y){]}^2}{M \times N}
\end{equation}
where $M$ and $N$ are the total number of pixels of image $L$ and image $L'$ respectively.
Generally, a higher PSNR value indicates greater similarity between the images.

\subsubsection{Distance reciprocal distortion (DRD)}
${\rm DRD}$ represents a measure of visual distortion in the binary images, which is shown in the following equation:
\begin{equation}
\label{eq:DRD}
{\rm DRD} = \frac{{\rm \sum_k DRD_k}}{{\rm NUBN}}
\end{equation}
where ${\rm NUBN}$ corresponds to the number of uneven $8\times8$ blocks in the Ground-Truth images, and ${\rm DRD_k}$ is defined as:
\begin{equation}
{\rm DRD_k}=\sum_{i=-2}^{2} \sum_{j=-2}^{2} | {\rm G_k(i, j)-B_{k}(x, y) | \times N_{w}(i, j)}
\end{equation}
where ${\rm DRD_k}$ indicates the distortion of the ${\rm k_{th}}$ flipped pixel, calculated using a $5\times5$ normalized weight matrix ${\rm N_w}$. ${\rm B_k}$ refers to the pixel of the binary images, and ${\rm G_k}$ denotes the pixel of the Ground-Truth images.

\subsubsection{Avg-Score (Avg)}
The PSNR metric provides a brief and intuitive quantitative way to compare the difference between the generated images and the Ground-Truth images, which can be used as a supplementary metric but should not be used as the only evaluation metric.
The FM and p-FM metrics are used to assess the accuracy of different methods, but higher values of these metrics do not always correspond to higher PSNR values.

Based on the results of our extensive experiments, it is common to observe that a method achieves the SOTA PSNR value, while the FM and p-FM values are lower than those of other models.
To solve this problem, we follow the Avg-Score metric used by Jemni \emph{et al.}~\cite{jemni2022enhance} to compare the combined performance of different methods, as shown in the following equation:
\begin{equation}
Avg = \frac{FM + p\!-\!FM + PSNR + (100-DRD)}{4}
\end{equation}

\begin{table*}[ht]
\centering
\caption{Quantitative comparison of PSNR values (dB) of images processed by different techniques with the corresponding Ground-Truth images on different training sets.
The highest and 2nd highest PSNR values are in {\color{red}red} and {\color{blue}blue} colors, respectively.}
\label{tab:psnr}
\setlength{\tabcolsep}{1.8mm}{
\begin{tabular}{lccccccc}
\hline \noalign{\smallskip}
\textbf{Method} & \textbf{DIBCO 2009} & \textbf{H-DIBCO 2010} & \textbf{H-DIBCO 2012} & \textbf{Bickley Diary} & \textbf{PHIBD} & \textbf{SMADI} & \textbf{Mean Values} \\ \noalign{\smallskip} \hline \noalign{\smallskip}
Original & {\color{blue}65.9812} & 62.7183 & {\color{blue}63.6814} & 59.3267 & {\color{blue}65.0543} & {\color{blue}62.5198} & {\color{blue}63.2136} \\ 
\noalign{\smallskip} \hline \noalign{\smallskip}
LL (DWT) & 56.8869 & 62.9102 & 57.4896 & 50.4236 & 51.6242 & 57.1778 & 56.0854 \\
HL (DWT) & 49.8898 & 49.7108 & 49.9529 & 50.8580 & 50.0277 & 49.7025 & 50.0236 \\
LH (DWT) & 49.9061 & 49.7136 & 49.9578 & 50.8483 & 50.0311 & 49.6946 & 50.0253 \\
HH (DWT) & 49.0228 & 48.8979 & 49.2434 & 50.0180 & 49.2218 & 48.9594 & 49.2272 \\
\noalign{\smallskip} \hline \noalign{\smallskip}
LL (DWT\&Norm) & {\color{red}70.2888} & {\color{red}77.0135} & {\color{red}74.3312} & {\color{red}65.1956} & {\color{red}71.4992} & {\color{red}69.4219} & {\color{red}71.2917} \\
HL (DWT\&Norm) & 56.8078 & 58.2430 & 59.3027 & 56.6744 & 57.1121 & 56.7830 & 57.6497 \\
LH (DWT\&Norm) & 56.0911 & 58.7330 & 58.9662 & 57.0237 & 57.9381 & 56.9083 & 57.6101 \\
HH (DWT\&Norm) & 63.5701 & {\color{blue}64.5478} & 63.4654 & {\color{blue}61.8377} & 61.6075 & 61.7661 & 62.7991 \\ 
\noalign{\smallskip} \hline
\end{tabular}}
\end{table*}

\begin{table*}[ht]
\centering
\caption{Quantitative comparison (FM/p-FM/PSNR/DRD/Avg) of the proposed methods with different patch sizes ($224\times224$ vs. $256\times256$) on multiple benchmark datasets.}
\label{tab:imgsize}
\setlength{\tabcolsep}{0.8mm}{
\begin{tabular}{ccccccccc}
\hline \noalign{\smallskip}
 &  & \textbf{DIBCO 2011} & \textbf{DIBCO 2013} & \textbf{H-DIBCO 2014} & \textbf{H-DIBCO 2016} & \textbf{DIBCO 2017} & \textbf{H-DIBCO 2018} & \textbf{Mean Values} \\ \noalign{\smallskip} \hline \noalign{\smallskip}
\multirow{5}{*}{$256\times256$} & FM↑ & 94.11 & 95.06 & 96.62 & 91.40 & 90.95 & 91.76 & 93.32 \\
 & p-FM↑ & 96.97 & 97.30 & 98.01 & 96.22 & 93.97 & 95.74 & 96.37 \\
 & PSNR↑ & 20.48 & 22.16 & 22.25 & 19.63 & 18.55 & 19.98 & 20.51 \\
 & DRD↓ & 1.76 & 1.66 & 0.96 & 2.96 & 2.95 & 2.75 & 2.17 \\
 & Avg↑ & 77.45 & 78.22 & 78.98 & 76.07 & 75.13 & 76.18 & 77.01 \\ \noalign{\smallskip} \hline \noalign{\smallskip}
\multirow{5}{*}{$224\times224$} & FM↑ & 94.22 & 94.65 & 96.60 & 91.76 & 91.31 & 92.89 & 93.57 \\
 & p-FM↑ & 97.47 & 97.10 & 98.33 & 96.82 & 94.30 & 96.96 & 96.83 \\
 & PSNR↑ & 20.54 & 22.01 & 22.23 & 19.73 & 18.63 & 20.39 & 20.59 \\
 & DRD↓ & 1.69 & 1.98 & 0.97 & 2.81 & 2.88 & 2.23 & 2.09 \\
 & Avg↑ & 77.64 & 77.95 & 79.05 & 76.38 & 75.34 & 77.00 & 77.22 \\ \noalign{\smallskip} \hline
\end{tabular}}
\end{table*}

\begin{figure*}[ht]
\centering
\includegraphics[width=\linewidth]{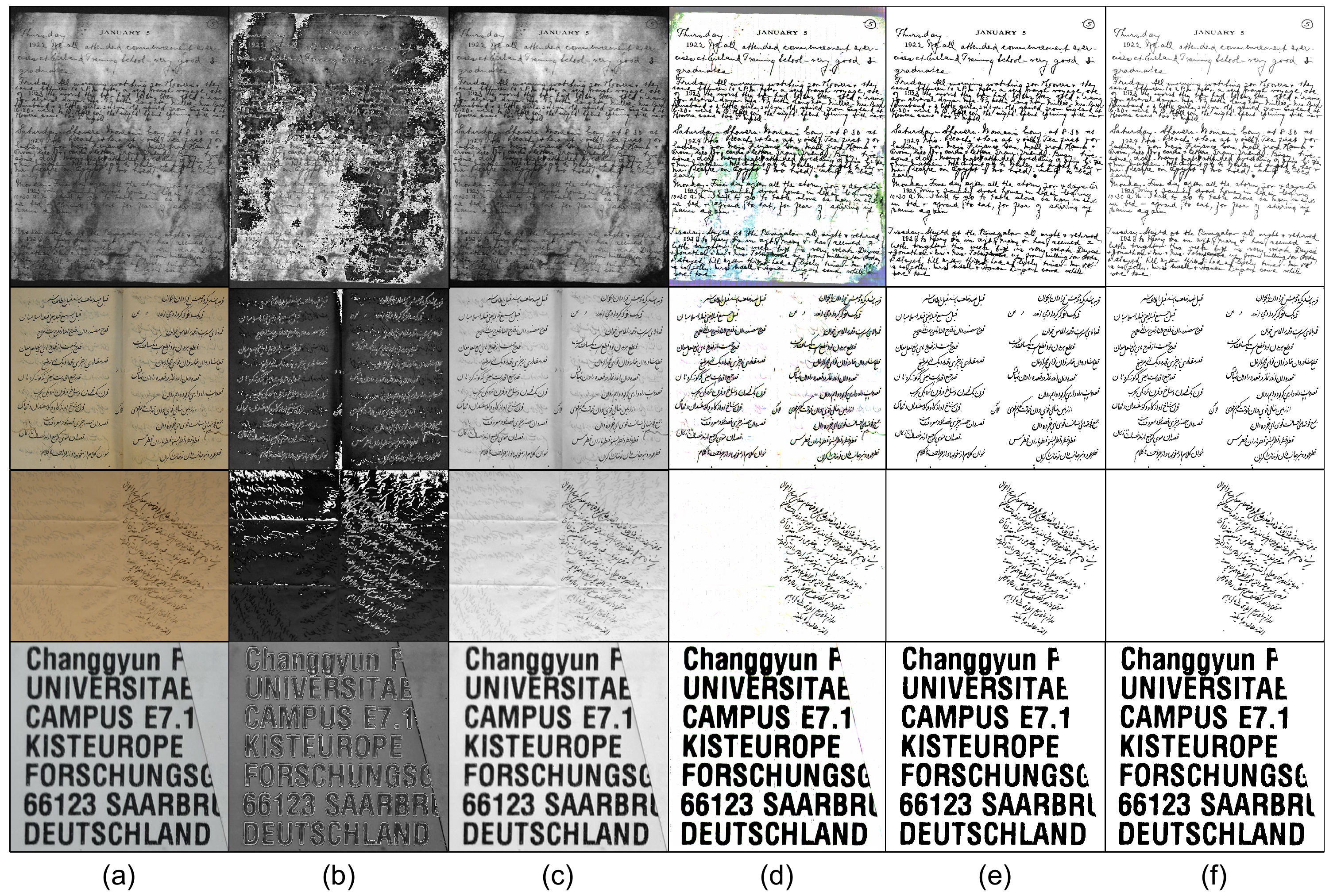}
\caption{Visualization examples of the results of the proposed method at each stage: (a) the original input images, (b) the LL subband images processed by Haar wavelet transform, (c) the LL subband images processed by Haar wavelet transform with normalization (Stage-1), (d) the enhanced images produced using our image enhancement method (Stage-2), (e) the binarized images obtained by combining local and global predictions (Stage-3), (f) the Ground-Truth images.}
\label{figure_enhancement&binarization}
\end{figure*}

\begin{table*}[ht]
\centering
\caption{Quantitative comparison of performance evaluation (FM/p-FM/PSNR/DRD/Avg) of the proposed method and other methods for document image binarization on multiple benchmark datasets.
Best and 2nd best performance are in {\color{red}red} and {\color{blue}blue} colors, respectively.
The data for the experimental results of other methods are obtained from~\cite{suh2022two,yang2024gdb}.}
\subfigure[DIBCO 2011]{
\resizebox{\columnwidth}{!}{
\begin{tabular}{lccccc}
\hline \noalign{\smallskip}
\textbf{Method} & \textbf{FM↑} & \textbf{p-FM↑} & \textbf{PSNR↑} & \textbf{DRD↓} & \textbf{Avg↑} \\ \noalign{\smallskip} \hline \noalign{\smallskip}
Niblack \cite{niblack1985introduction} & 70.44 & 73.03 & 12.39 & 24.95 & 57.73 \\
Otsu \cite{otsu1979threshold} & 78.60 & 80.53 & 15.31 & 22.57 & 62.97 \\
Sauvola \cite{sauvola2000adaptive} & 85.37 & 89.08 & 16.37 & 7.08 & 70.94 \\
Howe \cite{howe2013document} & {\color{blue}94.04} & 95.06 & {\color{blue}20.43} & 2.10 & 76.86 \\
Jia \cite{jia2018degraded} & 93.17 & 95.32 & 19.44 & 2.29 & 76.41 \\
Vo \cite{vo2018binarization} & 92.58 & 94.67 & 19.16 & 2.38 & 76.01 \\
He \cite{he2019deepotsu} & 91.92 & 95.82 & 19.49 & 2.37 & 76.22 \\
Suh \cite{suh2022two} & 93.57 & {\color{blue}95.93} & 20.22 & {\color{blue}1.99} & {\color{blue}76.93} \\
\textbf{Ours} & {\color{red}94.22} & {\color{red}97.47} & {\color{red}20.54} & {\color{red}1.69} & {\color{red}77.64} \\ \noalign{\smallskip} \hline
\end{tabular}}}
\subfigure[DIBCO 2013]{
\resizebox{\columnwidth}{!}{
\begin{tabular}{lccccc}
\hline \noalign{\smallskip}
\textbf{Method} & \textbf{FM↑} & \textbf{p-FM↑} & \textbf{PSNR↑} & \textbf{DRD↓} & \textbf{Avg↑} \\ \noalign{\smallskip} \hline \noalign{\smallskip}
Niblack \cite{niblack1985introduction} & 71.38 & 73.17 & 13.54 & 23.10 & 58.75 \\
Otsu \cite{otsu1979threshold} & 80.04 & 83.43 & 16.63 & 10.98 & 67.28 \\
Sauvola \cite{sauvola2000adaptive}& 82.71 & 87.74 & 17.02 & 7.64 & 69.96 \\
Howe \cite{howe2013document} & 91.34 & 91.79 & 21.29 & 3.18 & 75.31 \\
Jia\cite{jia2018degraded} & 93.28 & 96.58 & 20.76 & 2.01 & 77.15 \\
Vo \cite{vo2018binarization} & 93.43 & 95.34 & 20.82 & 2.26 & 76.83 \\
He \cite{he2019deepotsu} & 93.36 & {\color{blue}96.70} & 20.88 & 2.15 & 77.20 \\
Suh \cite{suh2022two} & {\color{red}95.01} & 96.49 & {\color{blue}21.99} & {\color{red}1.76} & {\color{blue}77.93} \\
\textbf{Ours} & {\color{blue}94.65} & {\color{red}97.10} & {\color{red}22.01} & {\color{blue}1.98} & {\color{red}77.95} \\ \noalign{\smallskip} \hline
\end{tabular}}}
\vskip 0.1in
\subfigure[H-DIBCO 2014]{
\resizebox{\columnwidth}{!}{
\begin{tabular}{lccccc}
\hline \noalign{\smallskip}
\textbf{Method} & \textbf{FM↑} & \textbf{p-FM↑} & \textbf{PSNR↑} & \textbf{DRD↓} & \textbf{Avg↑} \\ \noalign{\smallskip} \hline \noalign{\smallskip}
Niblack \cite{niblack1985introduction} & 86.01 & 88.04 & 16.54 & 8.26 & 70.58 \\
Otsu \cite{otsu1979threshold} & 91.62 & 95.69 & 18.72 & 2.65 & 75.85 \\
Sauvola \cite{sauvola2000adaptive} & 84.70 & 87.88 & 17.81 & 4.77 & 71.41 \\
Howe \cite{howe2013document} & 96.49 & 97.38 & {\color{red}22.24} & 1.08 & 78.76 \\
Jia \cite{jia2018degraded} & 94.89 & 97.68 & 20.53 & 1.50 & 77.90 \\
Vo \cite{vo2018binarization} & 95.97 & 97.42 & 21.49 & 1.09 & 78.45 \\
He \cite{he2019deepotsu} & 95.95 & {\color{red}98.76} & 21.60 & 1.12 & {\color{blue}78.80} \\
Suh \cite{suh2022two} & {\color{blue}96.36} & 97.87 & 21.96 & {\color{blue}1.07} & 78.78 \\
\textbf{Ours} & {\color{red}96.60} & {\color{blue}98.33} & {\color{blue}22.23} & {\color{red}0.97} & {\color{red}79.05} \\ \noalign{\smallskip} \hline
\end{tabular}}}
\subfigure[H-DIBCO 2016]{
\resizebox{\columnwidth}{!}{
\begin{tabular}{lccccc}
\hline \noalign{\smallskip}
\textbf{Method} & \textbf{FM↑} & \textbf{p-FM↑} & \textbf{PSNR↑} & \textbf{DRD↓} & \textbf{Avg↑} \\ \noalign{\smallskip} \hline \noalign{\smallskip}
Niblack \cite{niblack1985introduction} & 72.57 & 73.51 & 13.26 & 24.65 & 58.67 \\
Otsu \cite{otsu1979threshold} & 86.59 & 89.92 & 17.79 & 5.58 & 72.18 \\
Sauvola \cite{sauvola2000adaptive} & 84.64 & 88.39 & 17.09 & 6.27 & 70.96 \\
Howe \cite{howe2013document} & 87.47 & 92.28 & 18.05 & 5.35 & 73.11 \\
Jia \cite{jia2018degraded} & 90.10 & 93.72 & 19.00 & 4.03 & 74.70 \\
Vo \cite{vo2018binarization} & 90.01 & 93.44 & 18.74 & 3.91 & 74.57 \\
He \cite{he2019deepotsu} & 91.19 & 95.74 & 19.51 & 3.02 & 75.86 \\
Suh \cite{suh2022two} & {\color{red}92.24} & {\color{blue}95.95} & {\color{red}19.93} & {\color{red}2.77} & {\color{blue}76.34} \\
\textbf{Ours} & {\color{blue}91.76} & {\color{red}96.82} & {\color{blue}19.73} & {\color{blue}2.81} & {\color{red}76.38} \\ \noalign{\smallskip} \hline
\end{tabular}}}
\vskip 0.1in
\subfigure[DIBCO 2017]{
\resizebox{\columnwidth}{!}{
\begin{tabular}{lccccc}
\hline \noalign{\smallskip}
\textbf{Method} & \textbf{FM↑} & \textbf{p-FM↑} & \textbf{PSNR↑} & \textbf{DRD↓} & \textbf{Avg↑} \\ \noalign{\smallskip} \hline \noalign{\smallskip}
Otsu \cite{otsu1979threshold} & 77.73 & 77.89 & 13.85 & 15.54 & 63.48 \\
Sauvola \cite{sauvola2000adaptive} & 77.11 & 84.10 & 14.25 & 8.85 & 66.65 \\
Howe \cite{howe2013document} & 90.10 & 90.95 & {\color{blue}18.52} & 5.12 & 73.61 \\
Jia \cite{jia2018degraded} & 85.66 & 88.30 & 16.40 & 7.67 & 70.67 \\
Suh \cite{suh2022two} & {\color{blue}90.95} & {\color{red}94.65} & 18.40 & {\color{blue}2.93} & {\color{blue}75.27} \\
\textbf{Ours} & {\color{red}91.31} & {\color{blue}94.30} & {\color{red}18.63} & {\color{red}2.88} & {\color{red}75.34} \\ \noalign{\smallskip} \hline
\end{tabular}}}
\subfigure[H-DIBCO 2018]{
\resizebox{\columnwidth}{!}{
\begin{tabular}{lccccc}
\hline \noalign{\smallskip}
\textbf{Method} & \textbf{FM↑} & \textbf{p-FM↑} & \textbf{PSNR↑} & \textbf{DRD↓} & \textbf{Avg↑} \\ \noalign{\smallskip} \hline \noalign{\smallskip}
Otsu \cite{otsu1979threshold} & 51.45 & 53.05 & 9.74 & 59.07 & 38.79 \\
Sauvola \cite{sauvola2000adaptive} & 67.81 & 74.08 & 13.78 & 17.69 & 59.50 \\
Howe \cite{howe2013document} & 80.84 & 82.85 & 16.67 & 11.96 & 67.10 \\
Jia \cite{jia2018degraded} & 76.05 & 80.36 & 16.90 & 8.13 & 66.30 \\
Suh \cite{suh2022two} & {\color{blue}91.86} & {\color{blue}96.25} & {\color{blue}20.03} & {\color{blue}2.60} & {\color{blue}76.39} \\
\textbf{Ours} & {\color{red}92.89} & {\color{red}96.96} & {\color{red}20.39} & {\color{red}2.23} & {\color{red}77.00} \\ \noalign{\smallskip} \hline
\end{tabular}}}
\label{tab:result}
\end{table*}

\begin{figure*}[ht]
\centering
\includegraphics[width=\linewidth]{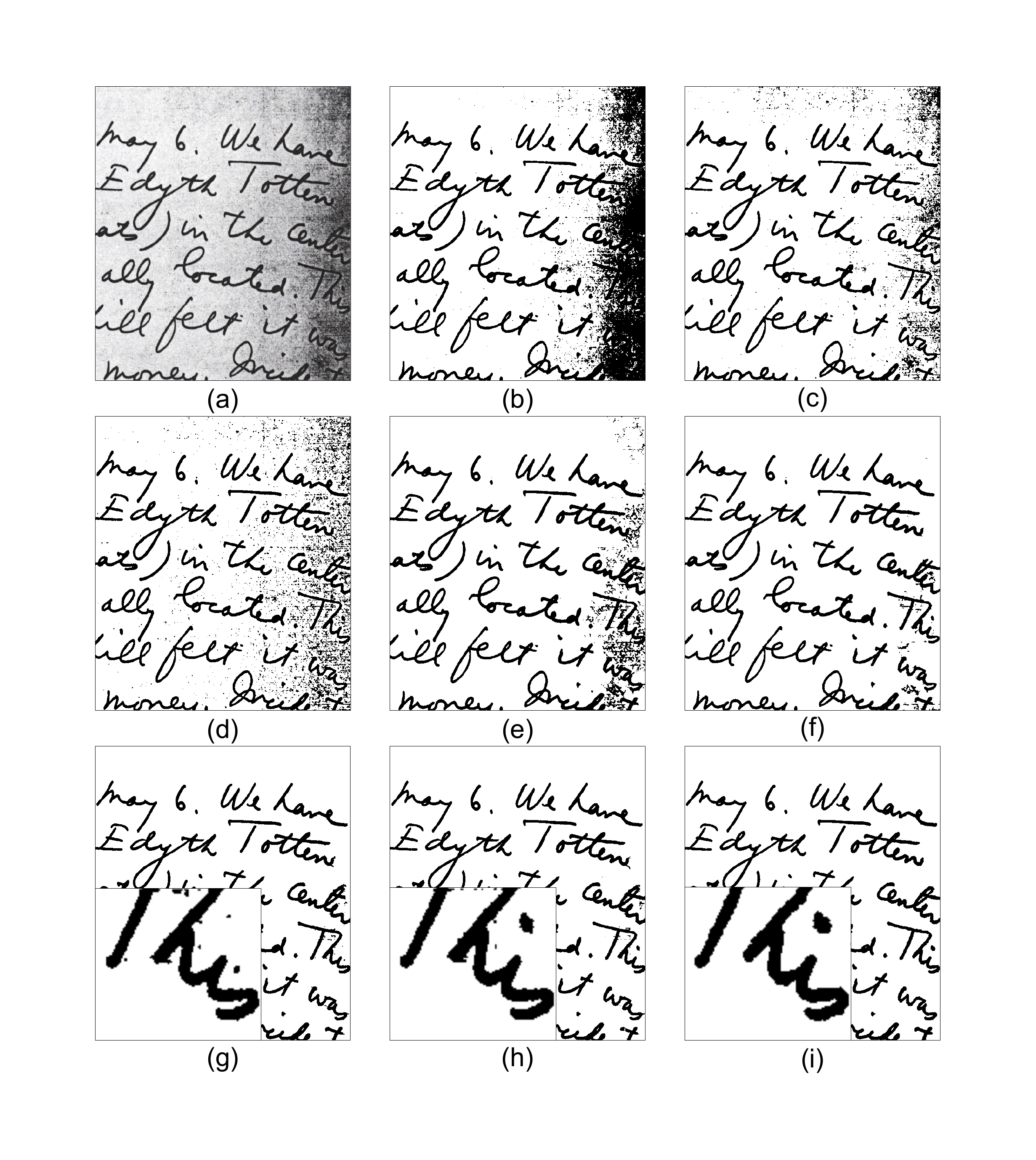}
\caption{Example of document image binarization results from the input image ``HW1'' of DIBCO 2011 dataset by different methods: (a) original input image, (b) Otsu's~\cite{otsu1979threshold}, (c) Niblack's~\cite{niblack1985introduction}, (d) Sauvola \emph{et al.}'s~\cite{sauvola2000adaptive}, (e) Vo \emph{et al.}'s~\cite{vo2018binarization}, (f) He \emph{et al.}'s~\cite{he2019deepotsu}, (g) Suh \emph{et al.}'s~\cite{suh2022two}, (h) Ours, and (i) Ground-Truth image.}
\label{figure_hw1}
\end{figure*}

\begin{figure*}[ht]
\centering
\includegraphics[width=\linewidth]{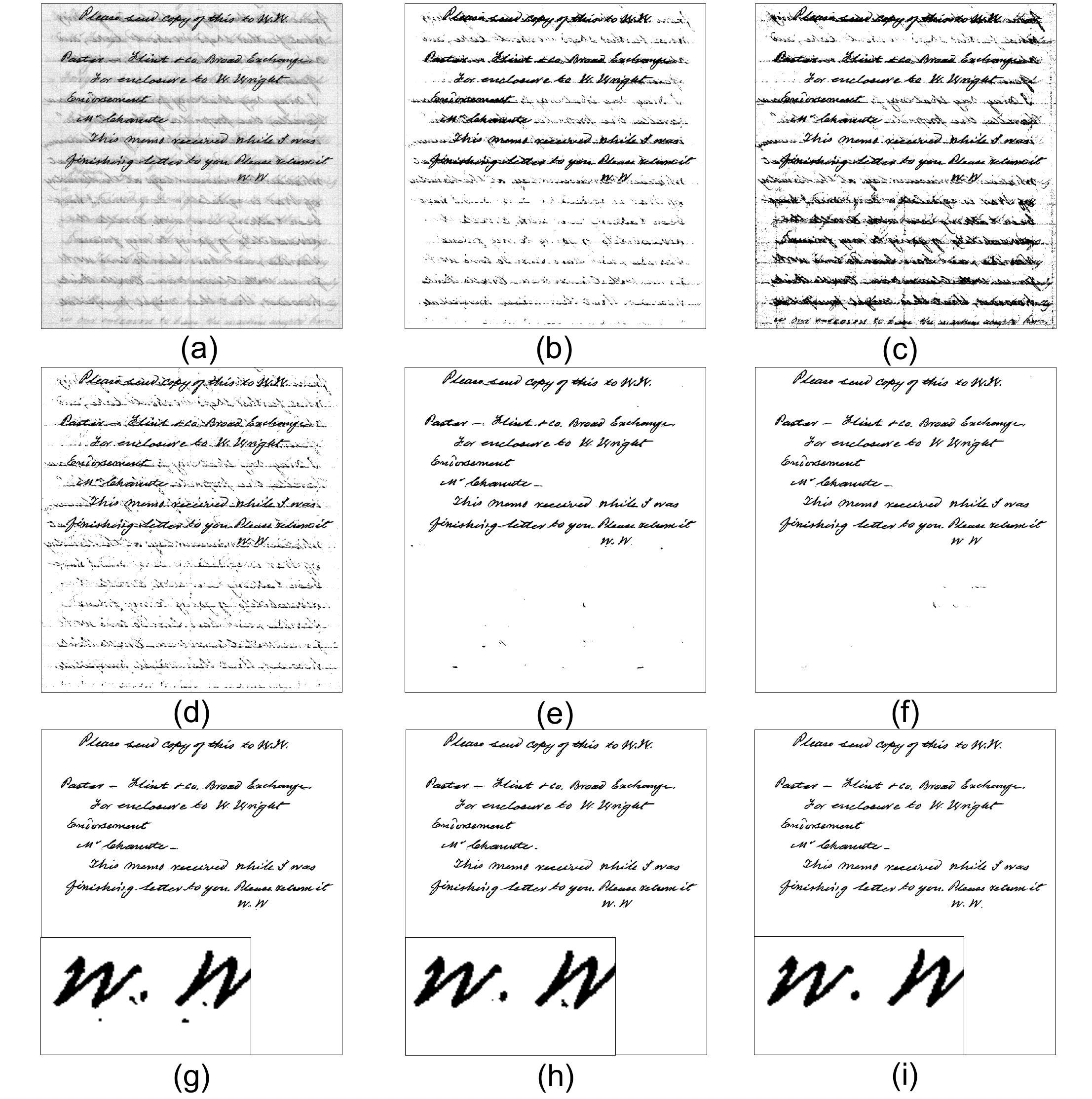}
\caption{Example of document image binarization results from input image ``HW5'' of DIBCO 2013 dataset by different methods: (a) original input image, (b) Otsu's~\cite{otsu1979threshold}, (c) Niblack's~\cite{niblack1985introduction}, (d) Sauvola \emph{et al.}'s~\cite{sauvola2000adaptive}, (e) Vo \emph{et al.}'s~\cite{vo2018binarization}, (f) He \emph{et al.}'s~\cite{he2019deepotsu}, (g) Suh \emph{et al.}'s~\cite{suh2022two}, (h) Ours, and (i) Ground-Truth image.}
\label{figure_hw5}
\end{figure*}

\subsection{Implementation Details}
\label{sec:implementation}
\subsubsection{Data preparation}
\label{subsec:data}
The model performance of the neural network depends on the quality and quantity of its training data.
In general, the more training data improves the better model performance.
To fairly compare the performance of the proposed method with other methods, we follow the strategy described in~\cite{vo2018binarization,he2019deepotsu,suh2022two} to construct the training and test sets.
Specifically, the training set includes a total of 143 images from DIBCO 2009~\cite{gatos2009icdar}, H-DIBCO 2010~\cite{pratikakis2010h}, H-DIBCO 2012~\cite{pratikakis2012icfhr}, PHIBD~\cite{ayatollahi2013persian}, SMADI~\cite{hedjam2013historical}, and BD~\cite{deng2010binarizationshop} dataset.
While the test sets comprise a total of 82 images from DIBCO 2011~\cite{pratikakis2011icdar2011}, DIBCO 2013~\cite{pratikakis2013icdar}, H-DIBCO 2014~\cite{ntirogiannis2014icfhr2014}, H-DIBCO 2016~\cite{pratikakis2016icfhr2016}, DIBCO 2017~\cite{pratikakis2017icdar2017}, and H-DIBCO 2018~\cite{pratikakis2018icdar2018}.

To ensure a fair comparison of the performance of all methods, for the proposed method and other methods, we apply the same data augmentation techniques to the training data.
In the training process of the GANs models of Stage-2, we split the original input images into patches of $224 \times 224$ size, and these patches are sampled with scale factors of ${0.75, 1, 1.25, 1.5}$.
In addition, we rotate the image by $270\degree$ to obtain more training data.
In contrast, for training GAN model on resized $512 \times 512$ images of Stage-3, we employ scaling without rotation for data augmentation.
To extend the training data, we employ the rotation augmentation technique during the training process of this model, including $90\degree$, $180\degree$ and $270\degree$ rotations, horizontal and vertical flips.

\subsubsection{Training}
The experiments are conducted using the PyTorch framework in Python.
We use single NVIDIA RTX 4090 GPU for all the experiments.

In this work, we employ the generator consisting of EfficientNet-B6~\cite{tan2019efficientnet} with U-Net++~\cite{zhou2018unet++,zhou2019unet++}.
About the training hyper-parameters, the settings of Stage 2 and Stage 3 are similar, with the only difference in the number of training epochs.
Specifically, for the network of the global prediction we set it to 150 epochs and for the remaining networks we set it to 10 epochs.
We select the Adam~\cite{kingma2014adam} optimizer, with an initial learning rate set to $2\times 10^{-4}$.
In addition, we set $\beta_1=0.5$ for the generator, and $\beta_2=0.999$ for the discriminator.

\subsubsection{Pre-training}
Due to the limited data available, to improve the training efficiency, we choose EfficientNet-B6~\cite{tan2019efficientnet} as the encoder of the generator, utilizing the pre-trained weights on the ImageNet dataset~\cite{deng2009imagenet}.

\subsection{Comparison with other Methods}
In our previous conference paper~\cite{ju2023ccdwt}, we compared different deep learning-based methods and found that some of these methods~\cite{tensmeyer2017document,bhunia2019improving,zhao2019document} employ ``leave-one-out'' strategy for constructing the training set.
For instance, when the DIBCO 2013~\cite{pratikakis2013icdar} dataset is used as the test set, all other DIBCO datasets are used as the training set.
Although our method still shows excellent performance in such case of~\cite{ju2023ccdwt}, we consider it is unfair to compare them directly because our model uses a much smaller training set.
Therefore, we exclude these methods from the methods we compare in this paper, and retain only those methods~\cite{niblack1985introduction,otsu1979threshold,sauvola2000adaptive,howe2013document,jia2018degraded,vo2018binarization,he2019deepotsu,suh2022two} that use the same training set for comparison.

\subsection{Image Enhancement Experiment}
\label{subsec:experiment_norm}
The previous conference paper~\cite{ju2023ccdwt} of this work presents ablation study on the DWT with normalization.
Based on this, it is better to intuitively show that, compared to the binarization results of the original input images, the binarization results of the images processed by the DWT with normalization are closer to the Ground-Truth images.

While the mathematical theory~\cite{stankovic2003haar} suggests that the images processed by the Haar wavelet transform can retain contour information and reduce noise more efficiently, it is important to demonstrate their impact on experimental results.
In Stage-1 of this work, we set the threshold to 0.5 to binarize different single-channel $224 \times 224$ patch images (red, green, and blue) to obtain the corresponding binarization results.
In the experiment of this section, we apply the above image binarization to the original input images, the images only processed by the DWT, and the images processed by the DWT with normalization.
Subsequently, we calculate the PSNR values of the binarized images by different techniques compared to the Ground-Truth images, and the results are presented in Table~\ref{tab:psnr}.
Compared with the PSNR mean value of the original input images of 63.2136dB, the PSNR mean values of the four subband images (LL, HL, LH, HH) only processed by the DWT are all lower, which are 56.0854dB, 50.0236dB, 50.0253dB, and 49.2272dB, respectively.
In addition, the PSNR mean values of the three subband images (HL, LH, LL) processed by the DWT with normalization are lower than that of the original input images, which are 57.6497dB, 57.6101dB, and 62.7991dB, respectively, while the PSNR mean value of the LL subband images processed by the DWT with normalization reaches 71.2917dB, which means they are closer to the Ground-Truth images.
The results of this experiment show that the choice of the DWT with normalization is more reasonable than only the DWT for processing the images.

\subsection{Comparison Experiment}
To ensure fairness in the experiments, we use the same training set and data augmentation techniques described in Section~\ref{subsec:data} for all the compared models.
Unlike Suh \emph{et al.}~\cite{suh2022two}, who split the input images into $256 \times 256$ patches, our proposed method sets the patch size to $224 \times 224$.
To demonstrate the positive impact of this change on the model performance, we conduct the experiments comparing models using $224 \times 224$ patches with those using $256 \times 256$ patches, keeping all other settings unchanged.
Table~\ref{tab:imgsize} presents the performance of the models across different patch sizes on six DIBCO datasets.
Since we have demonstrated in Section~\ref{subsec:experiment_norm} that the DWT with normalization technique outperforms the only DWT technique, the models in Table~\ref{tab:imgsize} all use the DWT with normalization technique.

According to Table~\ref{tab:imgsize}, the mean values of FM, p-FM, PSNR, DRD, and Avg for the model based on $224 \times 224$ patches are 93.57, 96.83, 20.59, 2.09, and 77.22, respectively, which are superior to those for the model based on $256 \times 256$ patches.
This indicates that the model based on $224 \times 224$ patches has better model performance, and therefore we will use the model based on $224 \times 224$ patches for subsequent comparisons.

\subsection{Experimental Results}
Different from the conference paper~\cite{ju2023ccdwt} of this work that uses the DWT with normalization to only process the input images, this paper uses the DWT with normalization to process both the input images and the Ground-Truth images of the three single-channel images (red, green, and blue) in Stage-1.

Figure~\ref{figure_enhancement&binarization} presents the visualization results at each stage of the proposed model.
Specifically, four images are randomly selected from the PHIBD~\cite{ayatollahi2013persian} and BD~\cite{deng2010binarizationshop} datasets to demonstrate the step-by-step process of document image enhancement and binarization.
Figure~\ref{figure_enhancement&binarization} (b) represents the results of the LL subband images procossed by the DWT, and (c) represents the results of the LL subband images procossed by the DWT with normalization.
It can be seen that there is a significant difference between (b) and (c), which performs noise reduction on the original input images.
Figure~\ref{figure_enhancement&binarization} (d) shows the effect of image enhancement achieved using the GANs models, where the images remove the background color and highlight the color of the text.
Figure~\ref{figure_enhancement&binarization} (e) presents the final output images obtained using the whole proposed method.
It is worth noting that the final outputs are very close to Figure~\ref{figure_enhancement&binarization} (f) the Ground-Truth images.

Table~\ref{tab:result} presents the evaluation results of different models on six DIBCO datasets.
Overall, the proposed method achieves the top two best performance in terms of FM, p-FM, PSNR, and DRD, while its Avg value reaching the SOTA level across all six datasets.
Notably, the performances of the proposed method on the DIBCO 2011~\cite{pratikakis2011icdar2011} and H-DIBCO 2018~\cite{pratikakis2018icdar2018} datasets are particularly good, obtaining the best values for each of the evaluation metrics.
For instance, on the DIBCO 2011 dataset, the p-FM value of the proposed method is 97.47, significantly surpassing the second best value of 95.93.
Similarly, on the H-DIBCO 2018 dataset, the FM value of the proposed method is 92.89, considerably higher than the second best value of 91.86.
These results indicate that the proposed method is more effective than other methods using the same training set for document image enhancement and binarization.

In addition to assessing method performance by comparing the quantitative metrics of different methods, this work aims to allow the reader to intuitively understand the differences between the methods by visualizing the binarization results.
Figures~\ref{figure_hw1} and \ref{figure_hw5} show the examples the binarization results by different methods for the image ``HW1'' of the DIBCO 2011 dataset, and ``HW5'' of the DIBCO 2013 dataset.
These figures visually demonstrate that the proposed method outperforms the traditional threshold methods and deep learning-based methods in terms of the shadow and noise elimination.
In addition, the proposed method retains the text content well while effectively mitigating the effects of shadows and noise.
Specifically, as shown in Figure~\ref{figure_hw5} (g) and (h), the proposed method is more efficient in eliminating noise compared to Suh \emph{et al.}'s~\cite{suh2022two} method.

\section{Conclusion and Future Work}
\label{sec:conclusion}
This paper introduces an effective three-stage network method for enhancement and binarization of gray-scale and color degraded document images.
Compared to other methods, the proposed method is more efficient in removing noise from degraded document images, and this effect applies to both gray-scale and color images.

As far as our knowledge extends, this work first represents the three-stage network method that combines traditional image processing with deep learning for document image enhancement and binarization.
The experimental results show that the performance of the proposed method using U-Net++ with EfficientNet-B6 as generator and PatchGAN as discriminator is better compared to other methods.
In addition, based on the experimental results, this paper also demonstrates that for document images, the binarization results of the images processed by the DWT with normalization are closer to the Ground-Truth images than the binarization results of the original input images.

We believe that this paper is of great significance to the reconstruction and analysis of ancient documents.
In future work, we will apply the proposed method to more degraded document images from different cultures to extract useful information to help culture organizations to analyze the history of ancient civilizations.
In particular, we will focus on the complex structure and polysemous characters, such as Chinese, Egyptian, Arabic, \emph{etc.}

\section{Declarations}
This paper is an expanded paper from 20th Pacific Rim International Conference on Artificial Intelligence (PRICAI) held on November 15-19, 2023 in Jakarta, Indonesia.

\subsection{Declaration of Funding}
This research is supported by National Science and Technology Council of Taiwan, under Grant Number: NSTC 112-2221-E-032-037-MY2.
\subsection{Declaration of competing interest}
The authors have no financial or proprietary interests in any material discussed in this article.
\subsection{Declaration of Generative AI and AI-assisted technologies in the writing process}
The authors only use generative artificial intelligence (AI) and AI-assisted technologies to improve the language.

\bibliographystyle{elsarticle-num}
\bibliography{cas-revised-refs}
\printcredits
\end{document}